\documentclass[letterpaper]{article} 
\usepackage{aaai2027}  
\nocopyright
\usepackage[hyphens]{url}  
\usepackage{graphicx} 
\usepackage{natbib}  
\usepackage{caption} 
\usepackage{algorithm}
\usepackage{algorithmic}

\usepackage{newfloat}
\usepackage{listings}
\DeclareCaptionStyle{ruled}{labelfont=normalfont,labelsep=colon,strut=off} 
\floatstyle{ruled}
\newfloat{listing}{tb}{lst}{}
\floatname{listing}{Listing}

\usepackage{booktabs}

\title{FlowVVTON: Flow-Guided Mask-Free Video Virtual Try-On}
\author{
    Shengyao Chen,
    Xianbing Sun,
    Liqing Zhang,
    Jianfu Zhang
}
\affiliations{
    Shanghai Jiao Tong University, Shanghai, China\\
    sjtu15060022751@sjtu.edu.cn, c.sis@sjtu.edu.cn
}

\begin{document}

\maketitle

\begin{abstract}
Video virtual try-on aims to transfer a target garment onto a moving person across video frames. Current methods rely on human parsing masks or pose keypoints that frequently fail under large motions and occlusions, causing boundary artifacts and temporal inconsistency. A further limitation is that most approaches rely solely on attention mechanisms for temporal modeling, providing no explicit motion supervision. We propose FlowVVTON, a mask-free framework that eliminates parsing mask dependency entirely. Optical flow is used solely as a training-time supervision signal: a flow-warped latent loss, applied across all layers of the generation model, enforces multi-scale temporal consistency by aligning adjacent-frame features under explicit physical motion constraints. A two-stage training strategy establishes mask-free spatial alignment before introducing flow-guided temporal supervision. Experiments on TikTokDress show that FlowVVTON outperforms baselines by substantial margins, particularly in temporal consistency (5.7$\times$ VFID-R improvement over SwiftTry), while requiring no segmentation masks, pose keypoints, or region annotations at any stage.
\end{abstract}

\begin{figure}[t]
  \centering
  \includegraphics[width=\columnwidth]{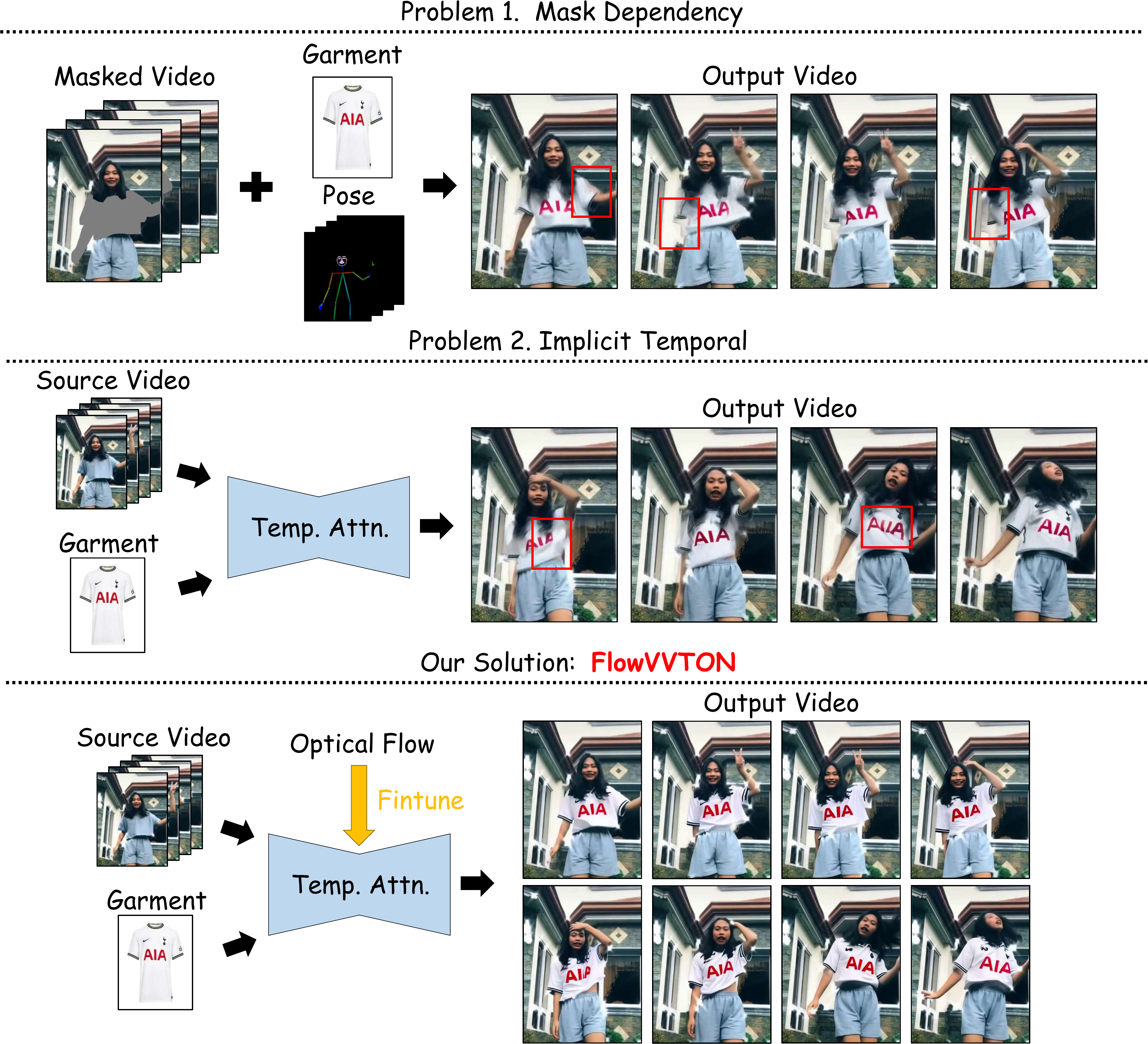}
  \caption{FlowVVTON performs mask-free video virtual try-on. Given only a source video and a target garment image, our method generates temporally consistent try-on results across all frames without any human parsing masks, pose keypoints, or region annotations. Temporal consistency is enforced during training via a flow-warped latent loss applied across all layers of the generation model.}
  \label{fig:teaser}
\end{figure}

\section{Introduction}

Video virtual try-on aims to realistically transfer a target garment onto a moving person across a video sequence. Unlike image-based try-on \cite{ddpm2020,ldm2022}, which operates on static frames, video try-on must satisfy \textit{temporal consistency}: the transferred garment must remain stable across frames without flickering, texture drift, or boundary artifacts, even as the person moves, rotates, or crosses their limbs. This capability has significant commercial value for online shopping, where dynamic garment visualization can reduce return rates, as well as for virtual content creation.

The dominant paradigm in video try-on conditions generation on human parsing masks or pose keypoints obtained from off-the-shelf estimators such as OpenPose \cite{openpose2021} or DensePose \cite{densepose2018}. These preprocessing tools segment the body into semantic regions (arms, torso, background) and provide spatial guidance for garment placement. However, they introduce a critical vulnerability: under fast motion, self-occlusion, or challenging lighting, mask and keypoint predictions become unreliable. Because mask extraction is the first stage of the pipeline, errors propagate and amplify through subsequent generation steps, manifesting as jagged garment edges, texture bleeding into the background, and temporal instability.

A second, related limitation is that most methods rely \textbf{solely} on attention mechanisms to maintain spatiotemporal consistency. They insert 3D convolutions or temporal attention layers into the denoising network and depend entirely on data-driven learning to capture cross-frame correspondence. While effective for slow, predictable motions, these attention layers provide no explicit motion supervision — the model must infer how pixels move between frames purely from statistical patterns in the training data. Under rapid movements or large body rotations, this implicit approach breaks down: without direct knowledge of pixel-level motion, textures drift, flicker, or deform.

We observe that both challenges share a common root: the absence of explicit, pixel-level motion information. \textbf{Optical flow} offers a natural solution. Optical flow is a dense, continuous field that describes precisely how each pixel moves from one frame to the next. Unlike discrete parsing masks, which classify each pixel into a fixed semantic category and are brittle to estimation errors, optical flow directly encodes the physical motion of every visible surface. This dual property — dense spatial correspondence and temporal motion encoding — makes optical flow a compelling alternative to masks for guiding video try-on. Our central question is therefore: \textit{can optical flow, used purely as a training-time supervision signal, enable effective and robust mask-free video try-on?}

We answer this question affirmatively with \textbf{FlowVVTON}, a mask-free video try-on framework. FlowVVTON takes only raw video frames and a garment image as input — no parsing masks, pose keypoints, or region annotations. The key innovation is a \textbf{flow-warped latent loss}: during training, optical flow is computed between adjacent frames and used to warp features across the temporal dimension at every layer of the generation model. The L2 distance between warped and original features is minimized, providing explicit, multi-scale supervision for temporal consistency. Crucially, optical flow is used only as a loss signal during training; at inference, the model generates temporally consistent video without any flow computation.

By removing mask dependency, FlowVVTON eliminates the dominant source of boundary artifacts in prior methods and handles large body motions more robustly. Training follows a two-stage pipeline: first, spatial alignment is established by fine-tuning on video frames; then, flow-guided temporal training enforces cross-frame consistency. We prepare synthetic paired training data from TikTokDress \cite{swifttry2025} using off-the-shelf video try-on models, and demonstrate through extensive experiments that FlowVVTON significantly outperforms existing baselines in both reconstruction quality and temporal consistency, achieving a 5.7$\times$ VFID-R improvement over SwiftTry.

Our contributions are:
\begin{enumerate}
    \item \textbf{A mask-free video try-on framework for large-motion scenarios.} FlowVVTON eliminates parsing mask dependency, avoiding mask-induced boundary artifacts and achieving robust performance under large body motions where mask-based methods typically fail. Optical flow is used only during training as a loss signal.
    \item \textbf{A per-layer flow-warped latent loss.} Applied across all layers of the generation model, the loss enforces multi-scale temporal consistency by aligning features between adjacent frames under physical motion.
    \item \textbf{A two-stage training strategy with flow-guided temporal supervision.} We design a training pipeline that first establishes mask-free spatial alignment on video frames, then enforces multi-scale temporal consistency through the flow-warped latent loss, enabling the model to handle large motions and reduce artifacts without any preprocessing masks.
\end{enumerate}

\section{Related Work}

\subsection{Image Virtual Try-On}

Image virtual try-on transfers a garment onto a person image. Early GAN-based methods \cite{viton2018,cpvton2018,clothflow2019,pfafn2021,vitonhd2021} used warp-and-blend pipelines effective under controlled conditions. Diffusion models \cite{ddpm2020,ldm2022} now dominate, using iterative denoising with garment feature injection via textual inversion, cross-attention fusion, or lightweight conditioning \cite{ladivton2023,stableviton2024,ootdiffusion2025,idmvton2024,catvton2024,leffa2025,siftvton2026}, with recent efforts targeting single-step generation \cite{pgvton2026,directtryon2026}. Mask-free image methods eliminate parsing dependency through pseudo-data training or unified architectures \cite{boowvton2025,mfviton2025,upvton2025,any2anytryon2025,dsvton2025}. We adopt a pretrained mask-free image model as our base.

\subsection{Video Virtual Try-On}

Video try-on adds temporal consistency: the garment must stay consistent across frames under motion. Early methods \cite{fwgan2019,clothformer2022} processed frames independently with post-hoc smoothing but lacked joint spatiotemporal modeling. Video diffusion models enabled end-to-end temporal generation, from SVD-based approaches with hierarchical attention and local refinement \cite{vivid2024,tunneltryon2024,wildvidfit2024,dpidm2025,realvvt2025,swifttry2025} to recent DiT-based architectures \cite{dit2023} with spatiotemporal encoding, stage-wise training, and keyframe-guided injection \cite{vitondit2025,catv2ton2025,magictryon2025,dreamvvt2025,keytailer2026,eevee2025,itryon2026}. All of these, however, rely on human parsing masks or pose keypoints \cite{openpose2021,densepose2018}.

\subsection{Mask-Free Try-On}

Mask-free methods avoid error propagation from imperfect upstream parsers. Image-domain works use pseudo-data training, teacher-student distillation, or unified architectures \cite{boowvton2025,mfviton2025,upvton2025,any2anytryon2025,dsvton2025,jcomvton2025}. Extending mask-free try-on to video is harder; recent methods replace dense masks with sparse keypoints, coarse bounding boxes, or pose-driven animation \cite{pemfvto2025,tripvvt2025,vanast2026} but still require spatial annotations or mask-based teachers during training. FlowVVTON eliminates masks entirely, using only raw video frames and a garment image as input, with flow-warped latent loss for temporal consistency.

\subsection{Optical Flow for Video Generation}

Optical flow has been used to improve video generation through flow-based losses \cite{flowloss2025,moalign2025}, as a conditioning signal via latent warping or noise correlation \cite{flowv2v2025,motionprompt2025,motionagent2025,flowvid2024,gowiththeflow2025}, or injected into model internals through frozen attention or warped embeddings \cite{animatediff2024,mogan2025,onlyflow2025,ropecraft2025}. In FlowVVTON, a per-layer flow-warped latent loss aligns adjacent-frame UNet features under motion constraints, providing explicit geometric supervision for temporal consistency.

\section{Method}

\begin{figure*}[t]
  \centering
  \includegraphics[width=\linewidth]{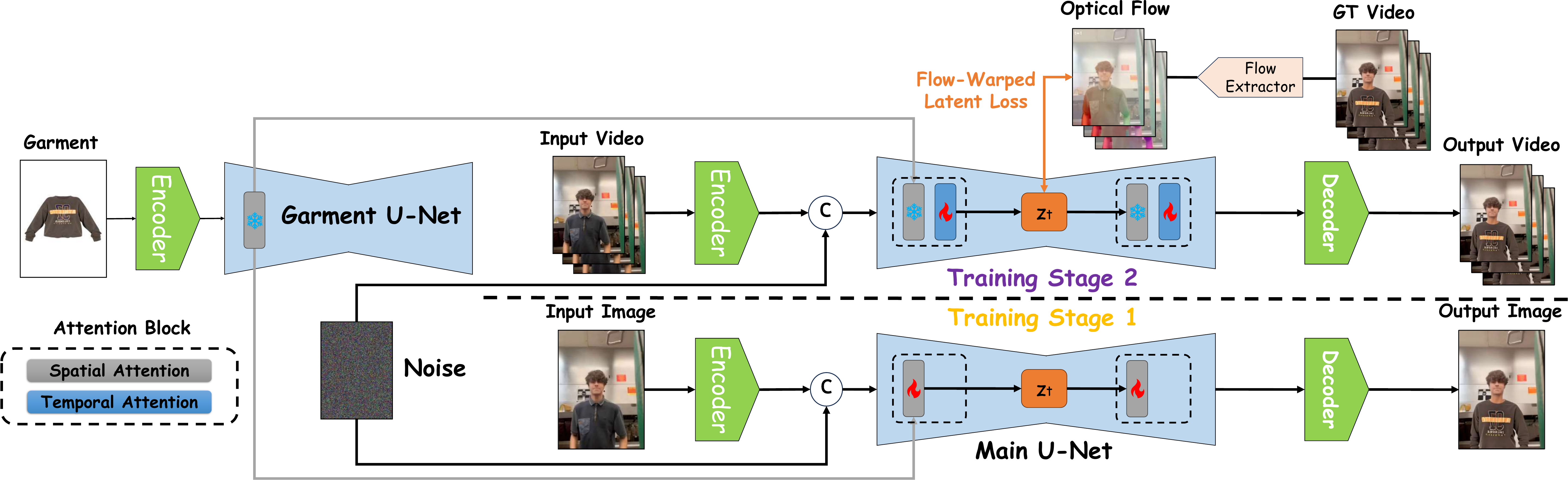}
  \caption{Overview of FlowVVTON. A reference UNet extracts garment features and injects them into the main UNet via mutual self-attention. The main UNet with temporal attention generates video frames from concatenated noisy and source latents. At every layer of the main UNet, a flow-warped latent loss enforces temporal consistency by aligning features across adjacent frames under optical flow guidance.}
  \label{fig:pipeline}
\end{figure*}

Our framework is built on a Latent Diffusion Model (LDM) \cite{ldm2022} backbone. All operations are performed in the latent space of a pretrained VAE, which compresses input video frames $V \in \mathbf{R}^{F \times H \times W \times 3}$ and the garment image $I_g$ into compact latent codes $z_v$ and $z_g$.

\subsection{Model Architecture}

\paragraph{Overview.}
The architecture consists of two UNets with shared base weights: a \textbf{reference UNet} for the garment image $I_g$, and a \textbf{main UNet} for video generation. The reference UNet extracts multi-scale garment features from $z_g$ and injects them into the main UNet via mutual self-attention: at each layer, the main UNet uses frame features as queries and garment features as keys and values. This WRITE-READ mechanism allows every spatial position to attend to garment texture without explicit warping.

The main UNet is a 2D UNet inflated to video by inserting temporal attention layers after each spatial block. Each residual block contains: (1) \textbf{spatial self-attention} with garment feature injection; and (2) \textbf{temporal self-attention} across $F$ frames to capture motion. Only temporal attention layers are trained; all spatial weights and the reference UNet remain frozen, preserving spatial quality from image pretraining.

\paragraph{Mask-free input.}
Instead of relying on parsing masks or pose keypoints, we feed raw video features directly. At each denoising step $t$, the main UNet receives the concatenation of the noisy latent $z_t$ and the VAE-encoded source video $z_{\mathrm{src}}$:
\begin{equation}
\mathrm{Input}_{\mathrm{main}} = \mathrm{Concat}(z_t, z_{\mathrm{src}}).
\end{equation}
$z_{\mathrm{src}}$ encodes the full original video — background, body contours, lighting, and motion — without any masking or region labeling. The model learns to distinguish garment areas from background purely through feature-level contrast, guided by strong semantic priors from large-scale image pretraining and garment features injected via mutual self-attention. This eliminates the error-prone mask extraction step entirely.

\paragraph{How optical flow is used.}
Optical flow serves two complementary roles during training, neither of which requires flow at inference. First, as a \emph{spatial alignment signal}: warping features across frames provides dense pixel-level correspondence, offering finer-grained supervision than the coarse region masks used by prior mask-based methods. Second, as a \emph{temporal regularizer}: the flow-warped consistency loss penalizes feature-level deviations between frames, forcing the model to learn temporally stable internal representations. Crucially, optical flow is only used during training as a loss signal; at inference, the model generates temporally consistent video without any flow computation.

\paragraph{Confidence-weighted warping.}
Optical flow is unreliable at occlusion boundaries and disoccluded regions where pixels appear or disappear between frames. Using erroneous flow as supervision would inject noise into training. We use forward-backward consistency checking: given $F_{t \to t+1}$ and $F_{t+1 \to t}$, we chain them and compute the forward-backward error $\mathrm{FBE}(p) = \|p - p''\|_2$. Pixels where forward and backward flows agree have low FBE and are reliable; large FBE indicates occlusion. We convert FBE into a confidence map $\mathbf{C}(p) = \exp(-\mathrm{FBE}(p) / \tau)$ with $\tau=3.0$, which down-weights unreliable pixels in the flow loss.

\paragraph{Why feature space?}
We apply warp consistency in UNet feature space rather than pixel space for three reasons. First, feature-space loss operates at the semantic level (texture, structure, garment identity), making it robust to lighting variations that would produce spurious pixel-space errors. Second, it directly supervises the features being trained rather than backpropagating through the VAE decoder. Third, features are available from the forward pass with no extra decoding cost.

\paragraph{Flow-warped latent loss.}
The loss operates within the main UNet on \emph{clean features}: during training, a clean forward pass (noise-free latents at timestep~0) yields multi-scale features $\mathbf{z}_l(t)$ for each frame $t$ and layer $l$, free from diffusion noise.

Given frames $x_{t}$ and $x_{t+1}$, we compute backward flow $F_{t+1 \to t}$ via RAFT \cite{raft2020} and downsample it to each UNet layer's resolution — coarse layers capture large body movements, fine layers track subtle texture displacements. The per-layer loss warps frame $t{+}1$ backward and measures MSE consistency with frame $t$:
\begin{equation}
\mathcal{L}_{l} = \frac{\sum_p \mathbf{C}_l(p) \cdot
\big\| \mathbf{z}_l(t,p) - \mathcal{W}\big(\mathbf{z}_l(t{+}1),\,F_{l}\big)(p) \big\|_2^2}
{\sum_p \mathbf{C}_l(p) + \varepsilon},
\label{eq:perlayer}
\end{equation}
where $\mathcal{W}$ is bilinear warping, $F_{l}$ is the flow downsampled to layer $l$, and $\mathbf{C}_l(p)$ is the confidence map. The per-layer losses are aggregated with level weights $w_l$ (2.0 for highest-resolution, 1.5 mid, 1.0 coarse) and per-layer variance normalization to prevent domination by high-activation layers.

For long-range stability, we compute the loss at multiple strides $s \in \{1, 2, 4\}$. Flow $F_{t+s \to t}$ is computed directly on sub-sampled frames, avoiding the compounding interpolation errors of accumulated stride-1 flows. The total flow loss is:
\begin{equation}
\mathcal{L}_{\mathrm{flow}} = \sum_{s \in \{1,2,4\}} \alpha_s \cdot
\sum_l w_l \, \mathcal{L}_{l}^{(s)},
\end{equation}
with $\alpha_1{=}1.0$, $\alpha_2{=}0.5$, $\alpha_4{=}0.25$. Stride-1 enforces adjacent-frame smoothness, stride-2 provides medium-range continuity, and stride-4 prevents long-term color drift. The decreasing weights reflect the decreasing reliability of flow over longer temporal intervals. By design, the flow-warped latent loss only involves the main UNet; the reference UNet is frozen and handles garment feature injection independently. This clean separation — spatial garment transfer via the frozen reference UNet, temporal consistency via flow-warped main UNet training — allows each component to be optimized without interference.

\begin{figure}[t]
  \centering
  \includegraphics[width=0.95\columnwidth]{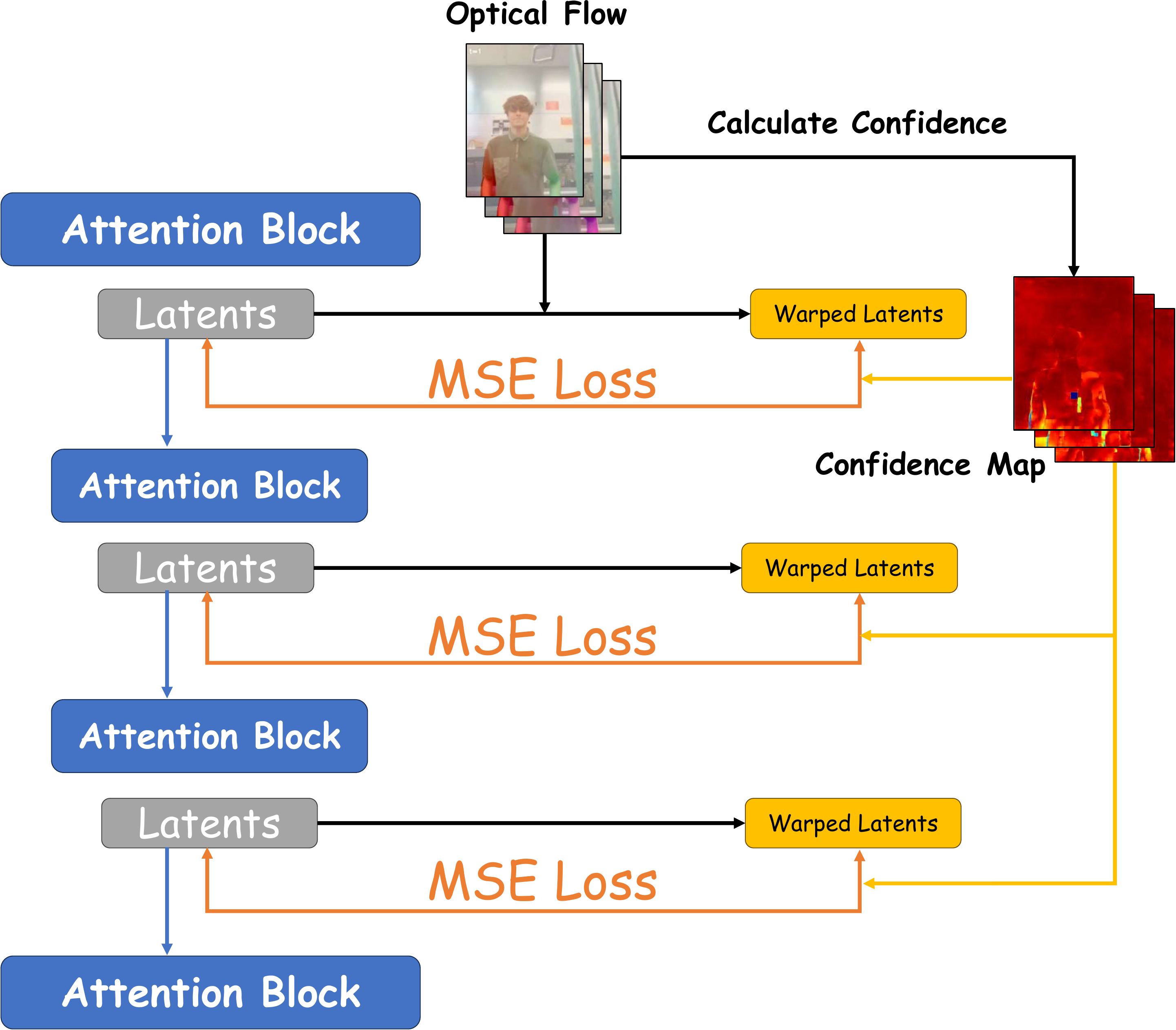}
  \caption{Detail of the flow-warped latent loss. Optical flow is computed between adjacent frames and downsampled to each layer. At every layer, the main UNet's clean features from frame $t{+}1$ are warped backward via optical flow and compared with frame $t$'s clean features via MSE loss. Confidence weights from forward-backward consistency suppress unreliable flow regions.}
  \label{fig:flow_loss_detail}
\end{figure}

\subsection{Data Preparation}

Paired video try-on data — the same person performing identical motion in different garments — is impractical to collect at scale. We prepare training data from TikTokDress \cite{swifttry2025} (693 training, 124 test videos), which contains diverse real-world motions, lighting, and backgrounds. An off-the-shelf video try-on model generates garment-transfer results for randomly sampled garments, producing large-scale video-garment pairs for model training. The synthesized videos serve as training inputs, while the original TikTokDress videos serve as ground truth for the denoising objective. An important practical benefit of this setup is that optical flow is computed on the clean ground-truth videos rather than on the synthesized results. Since GT videos are free from the minor artifacts and texture inconsistencies present in synthesized outputs, the resulting flow fields are substantially more reliable, providing cleaner motion supervision for the flow-warped latent loss.

\subsection{Training Strategy}

We adopt a two-stage training pipeline.

\paragraph{Stage 1: Image-level spatial alignment.}
We fine-tune a mask-free image try-on model on individual frames sampled from TikTokDress for 50K steps. This establishes per-frame garment transfer on diverse in-the-wild video frames, teaching the model to handle varied poses, lighting, and backgrounds without masks. Temporal attention layers remain frozen.

\begin{figure}[t]
  \centering
  \includegraphics[width=0.95\columnwidth]{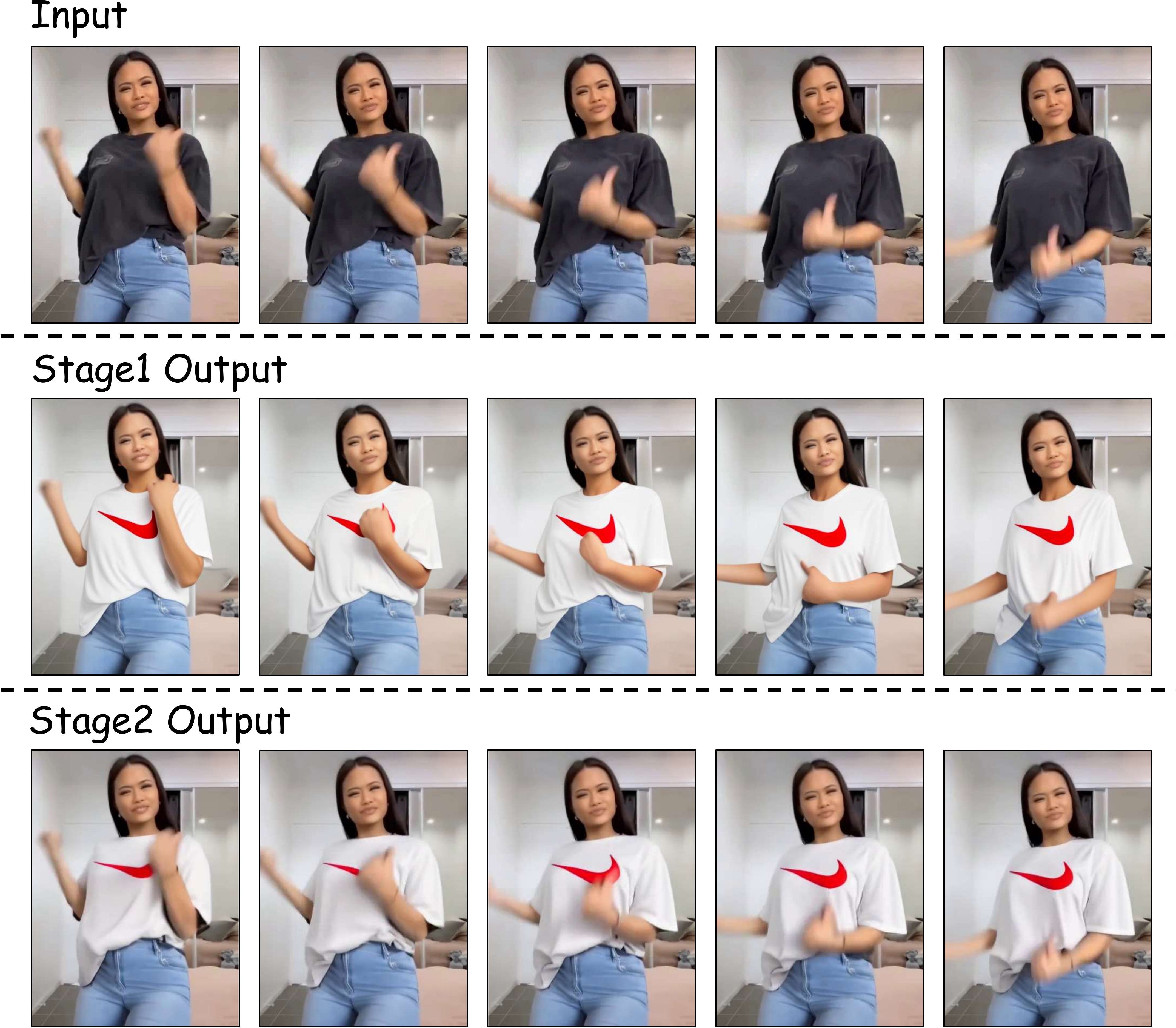}
  \caption{Stage 1 results. After image-level fine-tuning, the model achieves plausible per-frame garment transfer without masks, though temporal consistency is not yet enforced.}
  \label{fig:stage1}
\end{figure}

\paragraph{Stage 2: Video-level temporal training with flow guidance.}
We activate temporal attention layers and freeze all spatial weights from Stage~1. Training on our synthetic dataset introduces $\mathcal{L}_{\mathrm{flow}}$ across all main UNet layers. The temporal attention layers learn cross-frame motion patterns; the flow loss directly enforces geometric feature consistency between frames. This dual mechanism — implicit temporal attention for global coherence plus explicit flow constraints for local alignment — is the key to achieving stable mask-free video generation.

\paragraph{Training objective.}
The total loss combines standard diffusion denoising with the flow-warped latent loss:
\begin{equation}
\mathcal{L}_{\mathrm{total}} = \mathcal{L}_{\mathrm{ddpm}} + \lambda \mathcal{L}_{\mathrm{flow}},
\end{equation}
where $\mathcal{L}_{\mathrm{ddpm}}$ is the noise prediction MSE with Min-SNR weighting \cite{snr2023} ($\gamma=5.0$) and $\lambda$ controls the flow loss weight. Zero-terminal SNR \cite{zerosnr2023} is enabled for better noise scheduling at inference.

\section{Experiments}

\subsection{Datasets and Metrics}

\paragraph{Datasets.} We use the training data described in the Data Preparation section and evaluate under paired (same garment as source) and unpaired (different garment) settings on the TikTokDress test set.

\paragraph{Metrics.} For paired evaluation we report SSIM \cite{ssim} (structural similarity), LPIPS \cite{lpips} (perceptual distance), and VFID \cite{vfid} (Video Fr\'echet Inception Distance) with I3D and ResNeXt backbones. VFID$_{\mathrm{I3D}}$ is sensitive to spatiotemporal dynamics, while VFID$_{\mathrm{RN}}$ captures spatial texture quality. For unpaired evaluation we report VFID$_{\mathrm{I3D}}$ and VFID$_{\mathrm{RN}}$ as distribution-level measures. $\uparrow$ = higher is better, $\downarrow$ = lower is better.

\subsection{Implementation Details}

\paragraph{Architecture.} We use the pretrained VAE (ft-mse) from Stable Diffusion v1.5, frozen. The reference UNet is a frozen 2D UNet from a pretrained mask-free image try-on model. The main UNet is 3D-inflated with temporal attention layers after each spatial block; only these are trained.

\paragraph{Training.} Stage~1 fine-tunes an image try-on model on TikTokDress frames for 50K steps. Stage~2 trains temporal attention layers on our synthetic dataset for 100K steps, with AdamW (8-bit, lr=$1 \times 10^{-5}$, constant schedule), batch size 1, 16-frame sequences at stride~4, resolution $384 \times 512$, fp16 mixed precision, gradient clipping at 1.0, and gradient checkpointing.

\paragraph{Flow computation.} RAFT \cite{raft2020} with 32 update iterations computes flow on the fly. Forward-backward consistency filters unreliable regions ($\tau=3.0$). Flow loss weight $\lambda=1.0$, level weights: down\_0/up\_3: 2.0; down\_1/up\_2: 1.5; down\_2/up\_1: 1.0. Multi-scale strides $\{1,2,4\}$ with weights $\{1.0, 0.5, 0.25\}$.

\paragraph{Inference.} DDIM with 25 steps, CFG scale 3.5, context windows of 24 frames with 4-frame overlap, on a single NVIDIA A6000 GPU.

\subsection{Qualitative Comparison}

\begin{figure}[t]
  \centering
  \includegraphics[width=\columnwidth]{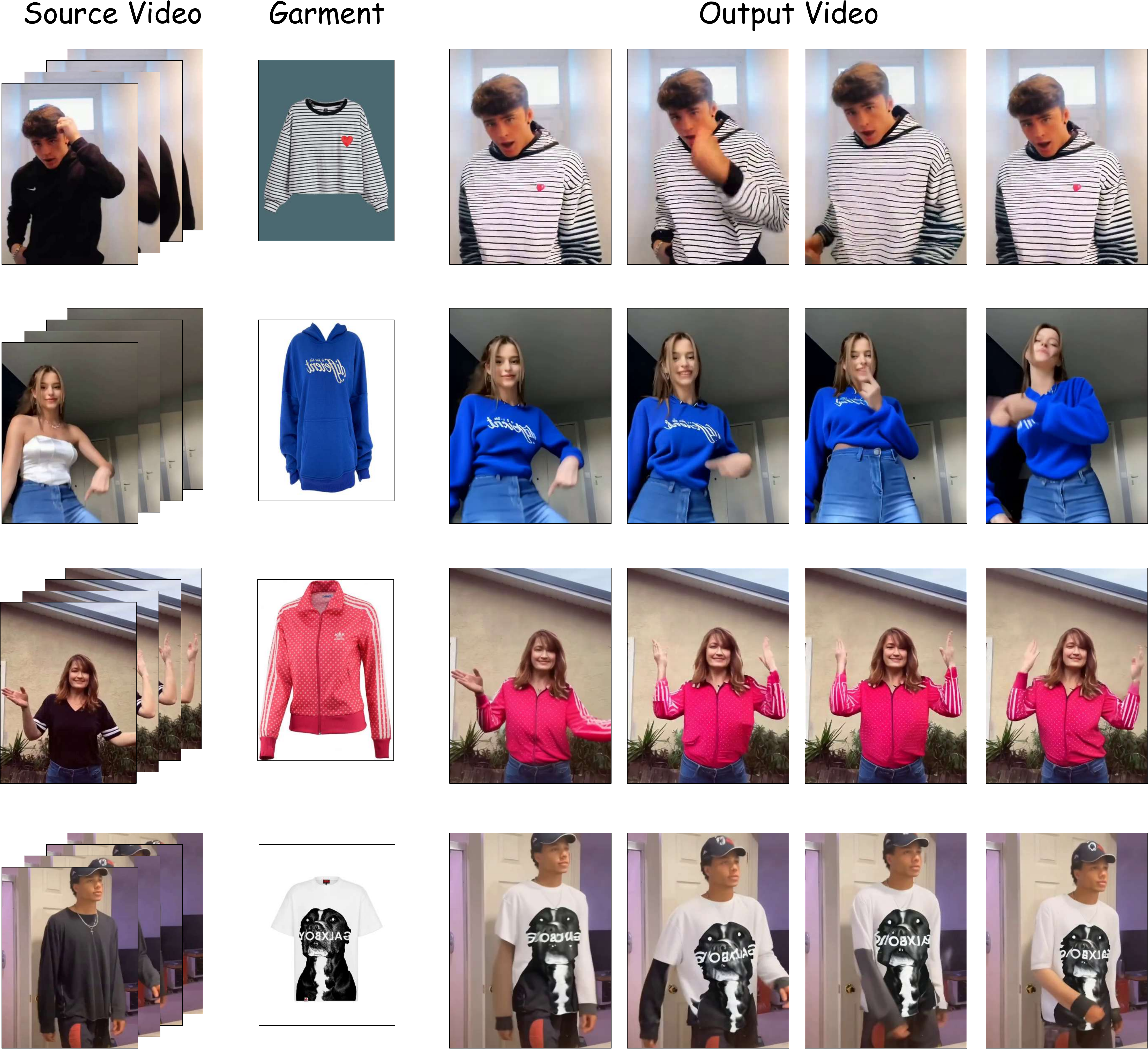}
  \caption{Qualitative results of FlowVVTON across diverse garments and motion patterns. Our method preserves garment textures and maintains temporal stability under large body motions, without mask-induced boundary artifacts.}
  \label{fig:qual_results}
\end{figure}

\begin{figure*}[t]
  \centering
  \includegraphics[width=\textwidth]{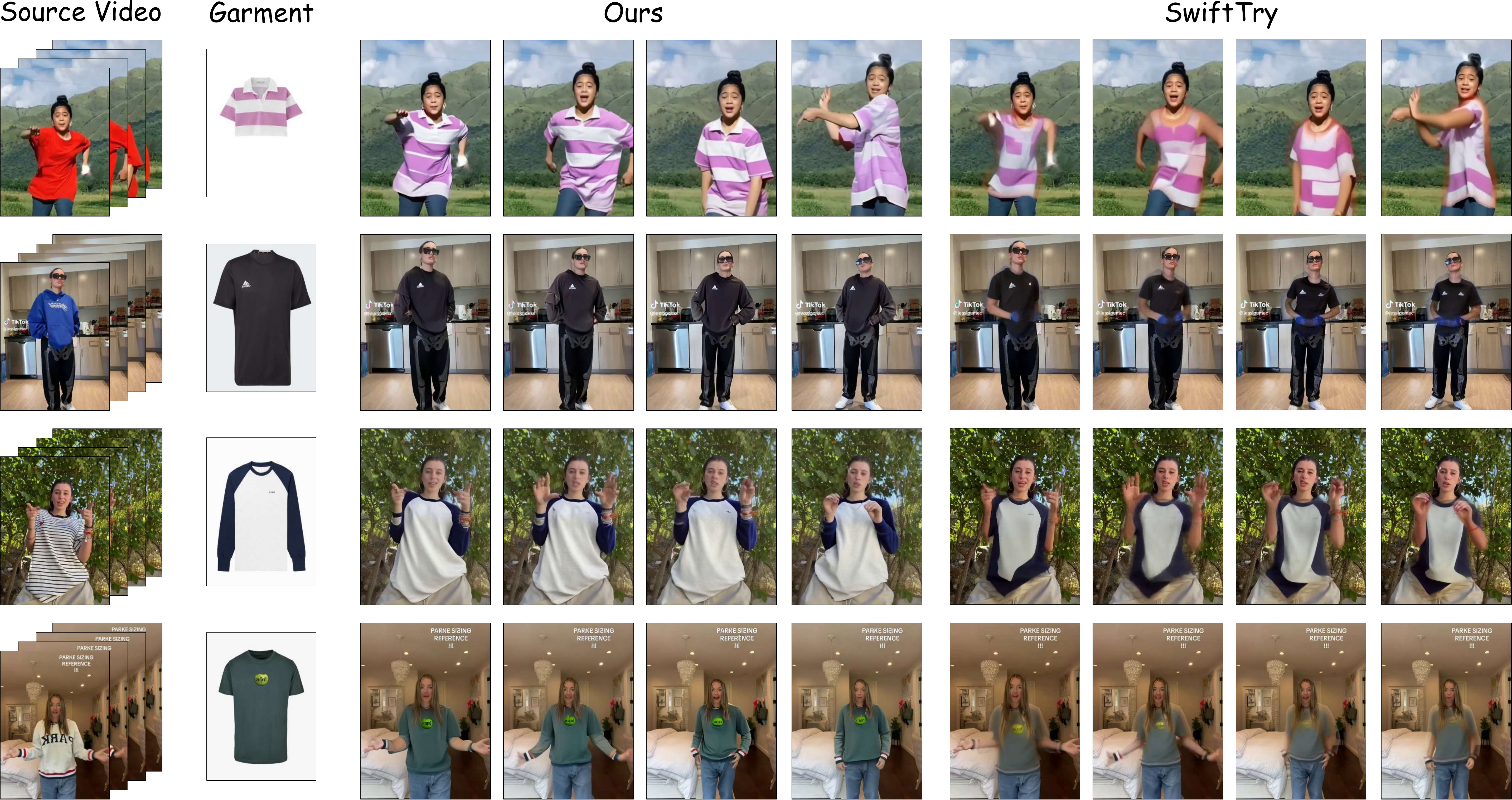}
  \caption{Qualitative comparison with SwiftTry \cite{swifttry2025}. FlowVVTON produces fewer boundary artifacts and sharper garment textures, particularly under large motions where mask-based methods degrade.}
  \label{fig:qual_comp}
\end{figure*}

Figure~\ref{fig:qual_comp} compares FlowVVTON with SwiftTry on representative test videos. FlowVVTON produces consistently sharper garment textures with significantly fewer boundary artifacts. The difference is most pronounced under large body motions: when the subject rotates rapidly or crosses their arms, SwiftTry's mask-based pipeline produces misaligned garment edges and visible texture bleeding, as the upstream parsing model fails to provide accurate spatial guidance. In contrast, FlowVVTON maintains clean garment boundaries and stable texture adherence in these challenging scenarios, because the mask-free input strategy preserves full visual context and the flow-warped latent loss provides continuous geometric alignment without relying on discrete, error-prone mask boundaries. Figure~\ref{fig:qual_results} shows additional results across diverse garment types and motion patterns, confirming that the mask-free design generalizes across varying conditions.

\subsection{Quantitative Comparison}

We evaluate under two settings: paired (same garment, same pose) and unpaired (different garment).

\begin{table}[t]
\centering
\setlength{\tabcolsep}{3.5pt}
\footnotesize
\caption{Paired evaluation on TikTokDress test set. $\uparrow$/$\downarrow$ = higher/lower is better. {\bf Bold} = best, \underline{underline} = second best.}
\label{tab:paired}
\begin{tabular}{lcccc}
\toprule
Method & SSIM$\uparrow$ & LPIPS$\downarrow$ & VFID-I$\downarrow$ & VFID-R$\downarrow$ \\
\midrule
SwiftTry & \underline{0.877} & \underline{0.090} & \underline{31.31} & \underline{4.30} \\
CatV2TON & 0.673 & 0.264 & 50.13 & 7.30 \\
MagicTryOn & 0.765 & 0.144 & 37.05 & 4.62 \\
\textbf{FlowVVTON (Ours)} & \textbf{0.905} & \textbf{0.072} & \textbf{29.94} & \textbf{0.71} \\
\bottomrule
\end{tabular}
\end{table}

\begin{table}[t]
\centering
\setlength{\tabcolsep}{8pt}
\small
\caption{Unpaired evaluation on TikTokDress test set. {\bf Bold} = best, \underline{underline} = second best.}
\label{tab:unpaired}
\begin{tabular}{lcc}
\toprule
Method & VFID-I$\downarrow$ & VFID-R$\downarrow$ \\
\midrule
SwiftTry           & \underline{32.06} & \underline{3.42} \\
CatV2TON           & 50.33 & 7.75 \\
MagicTryOn         & 34.46 & 4.02 \\
\textbf{FlowVVTON (Ours)} & \textbf{30.99} & \textbf{0.60} \\
\bottomrule
\end{tabular}
\end{table}

Table~\ref{tab:paired} reports paired metrics. FlowVVTON achieves the best results across all four, outperforming SwiftTry by substantial margins. The 6$\times$ VFID-R improvement (0.71 vs.\ 4.30) reflects the mask-free design's sharper garment details. In the unpaired setting (Table~\ref{tab:unpaired}), FlowVVTON again leads with a 5.7$\times$ VFID-R gain over SwiftTry (0.60 vs.\ 3.42). These consistent gains confirm that the flow-warped latent loss provides fundamentally better temporal supervision than implicit temporal attention. Notably, CatV2TON and MagicTryOn perform worse than SwiftTry, suggesting that DiT-based architectures do not compensate for the absence of explicit temporal constraints.

\subsection{Ablation Study}

\begin{table}[t]
\centering
\setlength{\tabcolsep}{8pt}
\small
\caption{Ablation study (unpaired evaluation). {\bf Bold} = best, \underline{underline} = second best.}
\label{tab:ablation}
\begin{tabular}{lcc}
\toprule
Configuration & VFID-I$\downarrow$ & VFID-R$\downarrow$ \\
\midrule
Full FlowVVTON (strides $\{1,2,4\}$) & 30.99 & \textbf{0.60} \\
\noalign{\vskip 2pt}\hline\noalign{\vskip 2pt}
\multicolumn{3}{l}{\textit{Loss components}}\\
\; w/o $\mathcal{L}_{\mathrm{flow}}$ ($\lambda{=}0$) & 32.48 & 1.24 \\
\; w/o confidence weighting & 31.99 & \underline{0.67} \\
\noalign{\vskip 2pt}\hline\noalign{\vskip 2pt}
\multicolumn{3}{l}{\textit{Multi-scale strides}}\\
\; w/o multi-scale (stride 1)   & \textbf{30.95} & 0.72 \\
\; strides $\{1,2\}$   & 31.19 & \underline{0.68} \\
\; strides $\{1,2,4\}$ & \underline{30.99} & \textbf{0.60} \\
\; strides $\{1,2,4,8\}$ & 31.03 & 0.67 \\
\bottomrule
\end{tabular}
\end{table}

\begin{figure}[t]
  \centering
  \includegraphics[width=0.95\columnwidth]{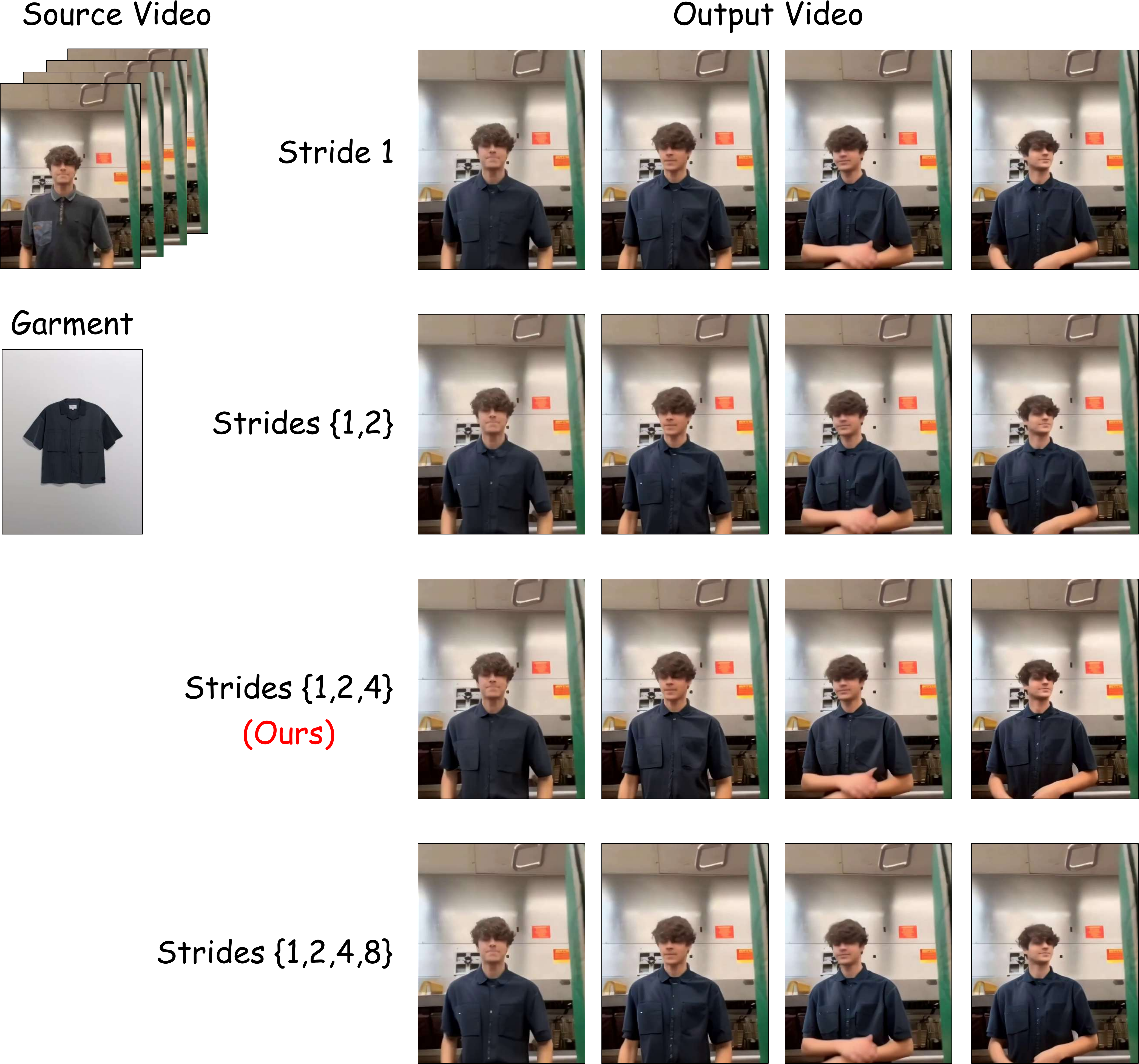}
  \caption{Qualitative ablation. Removing the flow loss causes severe texture drift; removing confidence weighting introduces artifacts near occlusions; stride-1 only produces local flickering. The full model with strides $\{1,2,4\}$ achieves the best temporal smoothness.}
  \label{fig:ablation_qual}
\end{figure}

Table~\ref{tab:ablation} and Figure~\ref{fig:ablation_qual} analyze each component's contribution. 

\paragraph{Loss components.} Removing the flow-warped latent loss ($\lambda{=}0$) causes the most severe degradation: VFID-I rises from 30.99 to 32.48 and VFID-R from 0.60 to 1.24, more than doubling the temporal error. This confirms that the standard denoising objective alone, even with temporal attention layers, cannot enforce adequate temporal consistency — explicit flow-based supervision is essential. Removing confidence weighting degrades VFID-R to 0.67, a smaller but notable drop; the qualitative results show artifacts concentrated near occlusion boundaries, confirming that the forward-backward consistency check effectively suppresses noisy flow in these regions. 

\paragraph{Multi-scale strides.} Using only stride-1 flow achieves the best VFID-I (30.95) but a substantially worse VFID-R (0.72), indicating that adjacent-frame smoothness alone is insufficient for temporal stability. Adding stride-2 improves VFID-R to 0.68 by providing intermediate-range constraints. The best configuration, strides $\{1,2,4\}$, achieves VFID-R of 0.60 — a 17\% improvement over stride-1 only — by combining short-range smoothness, mid-range alignment, and long-range temporal stability. Extending to stride-8 slightly degrades VFID-R to 0.67, as flow estimates over 8-frame intervals become less reliable due to accumulated motion changes and larger displacements. These results demonstrate that multi-scale flow supervision is critical for temporal consistency, and that the optimal stride range balances coverage against flow reliability.

\paragraph{Flow quality analysis.}
The quality of optical flow directly affects the effectiveness of the flow-warped latent loss. We ablate the number of RAFT update iterations.

\begin{table}[t]
\centering
\setlength{\tabcolsep}{8pt}
\small
\caption{RAFT iteration ablation (unpaired evaluation). {\bf Bold} = best.}
\label{tab:flow_quality}
\begin{tabular}{lcc}
\toprule
Configuration & VFID-I$\downarrow$ & VFID-R$\downarrow$ \\
\midrule
RAFT iters = 16 & 31.46 & 0.61 \\
RAFT iters = 32 (ours) & \underline{30.99} & \textbf{0.60} \\
RAFT iters = 48 & \textbf{30.92} & \textbf{0.60} \\
\bottomrule
\end{tabular}
\end{table}

\begin{figure}[t]
  \centering
  \includegraphics[width=0.95\columnwidth]{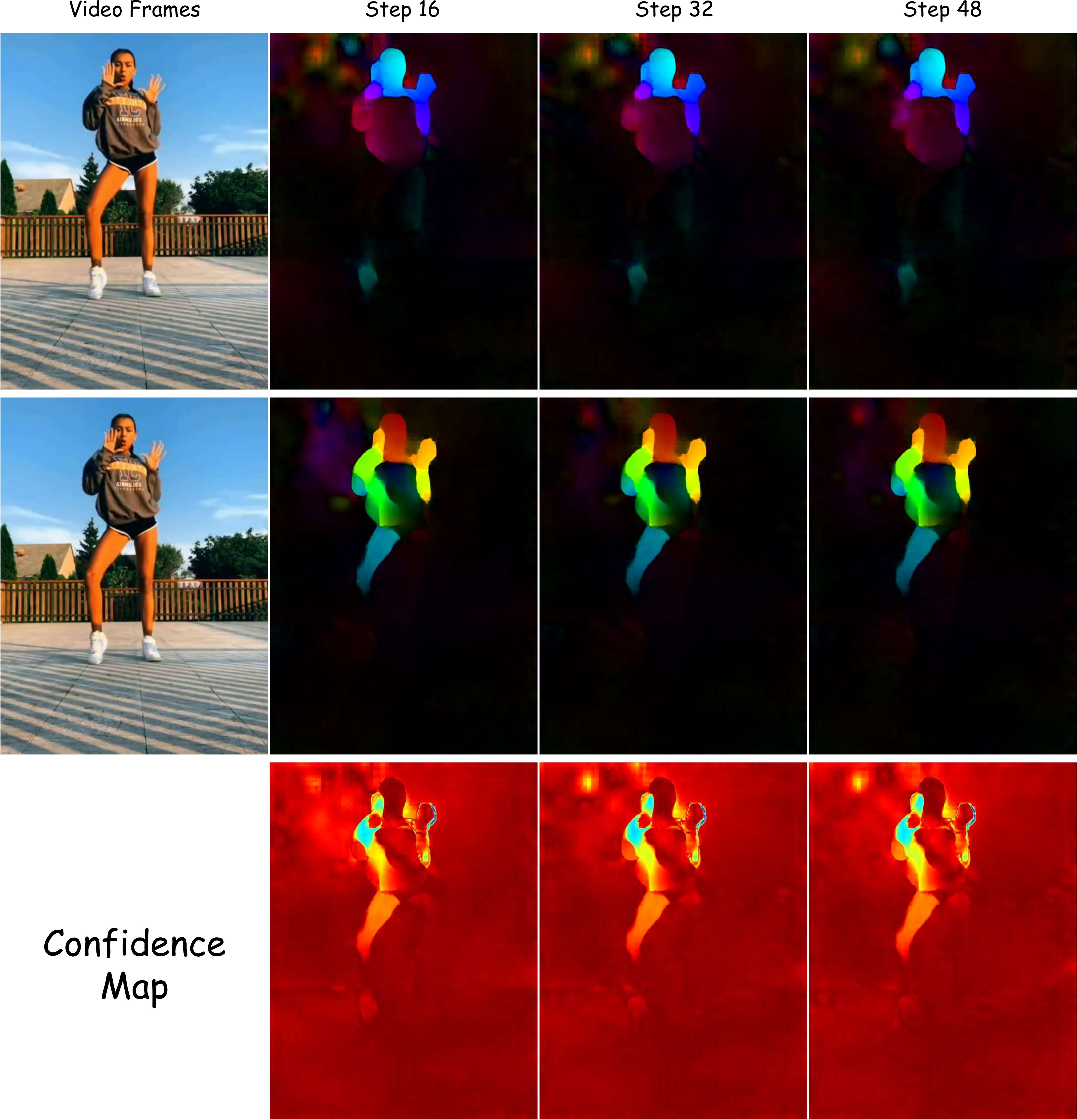}
  \caption{Flow quality analysis. The flow field at 16 iterations shows visible noise and misalignment at motion boundaries, while the 32-iteration result is substantially cleaner. The 48-iteration flow is nearly indistinguishable from 32, confirming diminishing returns.}
  \label{fig:flow_quality}
\end{figure}

Table~\ref{tab:flow_quality} and Figure~\ref{fig:flow_quality} analyze the sensitivity to optical flow quality. Reducing RAFT iterations from 32 to 16 causes a clear degradation across both metrics: VFID-I drops from 30.99 to 31.46 and VFID-R from 0.60 to 0.61. The qualitative results reveal the cause: at 16 iterations, the flow field exhibits visible noise and misalignment at motion boundaries, introducing erroneous supervision precisely where accurate alignment is most critical. Increasing to 48 iterations yields only a marginal VFID-I improvement (30.92) with identical VFID-R. The 48-iteration flow field is visually indistinguishable from 32, confirming diminishing returns. However, per-iteration training time increases measurably with more RAFT iterations, as each additional iteration requires another round of correlation lookup and GRU updates. We therefore select 32 iterations as the optimal operating point, where flow quality is sufficient for effective temporal supervision without unnecessary computational overhead.

\section{Conclusion}

We presented FlowVVTON, a mask-free video virtual try-on framework that eliminates preprocessing mask dependencies. By using optical flow solely as a training-time loss via a per-layer flow-warped latent loss, FlowVVTON achieves robust temporal consistency and reduces boundary artifacts, particularly under large motions where mask-based methods degrade. Our experiments demonstrate the flow-warped loss and multi-scale strides are essential.

Limitations include dependency on optical flow quality during training, UNet-specific architecture, and current resolution constraints. The mask-free paradigm and flow-guided temporal loss generalize to other video editing tasks such as video inpainting and motion-guided stylization.

\bibliography{aaai2027}

@inproceedings{viton2018,
  title     = {VITON: An Image-based Virtual Try-on Network},
  author    = {Han, Xintong and Wu, Zuxuan and Wu, Zhe and Yu, Ruichi and Davis, Larry S},
  booktitle = {CVPR},
  year      = {2018}
}

@inproceedings{cpvton2018,
  title     = {Toward Characteristic-Preserving Image-based Virtual Try-on Network},
  author    = {Wang, Bochao and Zheng, Huabin and Liang, Xiaodan and Chen, Yimin and Lin, Liang and Yang, Meng},
  booktitle = {ECCV},
  year      = {2018}
}

@inproceedings{clothflow2019,
  title     = {ClothFlow: A Flow-based Model for Clothed Person Generation},
  author    = {Han, Xintong and Hu, Xiaojun and Huang, Weilin and Scott, Matthew R},
  booktitle = {ICCV},
  year      = {2019}
}

@article{pfafn2021,
  title   = {Parser-Free Virtual Try-on via Distilling Appearance Flows},
  author  = {Ge, Yuying and Song, Yibing and Zhang, Ruimao and Ge, Chongjian and Liu, Wei and Luo, Ping},
  journal = {arXiv preprint arXiv:2103.04559},
  year    = {2021}
}

@inproceedings{vitonhd2021,
  title     = {VITON-HD: High-Resolution Virtual Try-On via Misalignment-Aware Normalization},
  author    = {Choi, Seunghwan and Park, Sunghyun and Lee, Minsoo and Choo, Jaegul},
  booktitle = {CVPR},
  year      = {2021}
}

@inproceedings{ladivton2023,
  title     = {Ladi-VTON: Latent Diffusion Textual-Inversion Enhanced Virtual Try-on},
  author    = {Morelli, Davide and Baldrati, Alberto and Cartella, Giuseppe and Cornia, Marcella and Bertini, Marco and Cucchiara, Rita},
  booktitle = {ACM Multimedia},
  year      = {2023}
}

@inproceedings{stableviton2024,
  title     = {StableVITON: Learning Semantic Correspondence with Latent Diffusion Model for Virtual Try-on},
  author    = {Kim, Jeongho and Gu, Guojung and Park, Minho and Park, Sunghyun and Choo, Jaegul},
  booktitle = {CVPR},
  year      = {2024}
}

@article{ootdiffusion2025,
  title   = {OOTDiffusion: Outfitting Fusion based Latent Diffusion for Controllable Virtual Try-on},
  author  = {Xu, Yuhao and Gu, Tao and Chen, Weifeng and Chen, Chengcai},
  journal = {arXiv preprint arXiv:2403.01779},
  year    = {2024}
}

@article{idmvton2024,
  title   = {Improving Diffusion Models for Authentic Virtual Try-on in the Wild},
  author  = {Choi, Yisol and Kwak, Sangkyung and Lee, Kyungmin and Choi, Hyungwon and Shin, Jinwoo},
  journal = {arXiv preprint arXiv:2403.05139},
  year    = {2024}
}

@inproceedings{catvton2024,
  title     = {CatVTON: Concatenation is All You Need for Virtual Try-on with Diffusion Models},
  author    = {Chong, Zheng and Dong, Xiao and Li, Haoxiang and Zhang, Wenqing and Zhao, Hanqing and Jiang, Dongmei and Liang, Xiaodan},
  booktitle = {ICLR},
  year      = {2025}
}

@inproceedings{leffa2025,
  title     = {Learning Flow Fields in Attention for Controllable Person Image Generation},
  author    = {Zhou, Zijian and Liu, Shikun and Han, Xiao and Liu, Haozhe and Ng, Kam Woh and Xie, Tian and Cong, Yuren and Li, Hang and Xu, Mengmeng and P{\'e}rez-R{\'u}a, Juan-Manuel},
  booktitle = {CVPR},
  year      = {2025}
}

@article{siftvton2026,
  title   = {SIFT-VTON: Geometric Correspondence Supervision on Cross-Attention for Virtual Try-On},
  author  = {Takemoto, Kosuke and Koshinaka, Takafumi},
  journal = {arXiv preprint arXiv:2605.01296},
  year    = {2026}
}

@article{pgvton2026,
  title     = {PG-VTON: A Novel Image-based Virtual Try-on Method via Progressive Inference Paradigm},
  author    = {Fang, Naiyu and Qiu, Lemiao and Zhang, Shuyou and Wang, Zili and Hu, Kerui},
  journal   = {IEEE Transactions on Multimedia},
  volume    = {26},
  year      = {2024},
  publisher = {IEEE}
}

@article{directtryon2026,
  title   = {DirectTryOn: One-Step Virtual Try-On via Straightened Conditional Transport},
  author  = {Sun, Xianbing and Zhan, Jiahui and Zhang, Liqing and Zhang, Jianfu},
  journal = {arXiv preprint arXiv:2605.12939},
  year    = {2026}
}

@inproceedings{boowvton2025,
  title     = {BooW-VTON: Boosting In-the-Wild Virtual Try-on via Mask-Free Pseudo Data Training},
  author    = {Zhang, Xuanpu and Song, Dan and Zhan, Pengxin and Chang, Tianyu and Zeng, Jianhao and Chen, Qingguo and Luo, Weihua and Liu, An-An},
  booktitle = {CVPR},
  year      = {2025}
}

@article{mfviton2025,
  title   = {MF-VITON: High-Fidelity Mask-Free Virtual Try-on with Minimal Input},
  author  = {Wan, Zhenchen and Hu, Dongting and Cheng, Weilun and Chen, Tianxi and Wang, Zhaoqing and Liu, Feng and Liu, Tongliang and Gong, Mingming},
  journal = {arXiv preprint arXiv:2503.08650},
  year    = {2025}
}

@inproceedings{upvton2025,
  title     = {UP-VTON: A Unified Virtual Try-On Framework Supporting Mask, Mask-Free, and Prompt-Driven Guidance},
  author    = {Jo, Youngjoo and Park, Minho and Kang, Dong-oh},
  booktitle = {ICCVW},
  year      = {2025}
}

@inproceedings{any2anytryon2025,
  title     = {Any2AnyTryOn: Leveraging Adaptive Position Embeddings for Versatile Virtual Clothing Tasks},
  author    = {Guo, Hailong and Zeng, Bohan and Song, Yiren and Zhang, Wentao and Liu, Jiaming and Zhang, Chuang},
  booktitle = {ICCV},
  year      = {2025}
}

@article{dsvton2025,
  title   = {DS-VTON: An Enhanced Dual-Scale Coarse-to-Fine Framework for Virtual Try-On},
  author  = {Sun, Xianbing and Hong, Yan and Zhan, Jiahui and Lan, Jun and Zhu, Huijia and Wang, Weiqiang and Zhang, Liqing and Zhang, Jianfu},
  journal = {arXiv preprint arXiv:2506.00908},
  year    = {2025}
}

@article{jcomvton2025,
  title   = {JCo-MVTON: Jointly Controllable Multi-Modal Diffusion Transformer for Mask-Free Virtual Try-on},
  author  = {Wang, Aowen and Li, Wei and Luo, Hao and Ao, Mengxing and Zhu, Chenyu and Li, Xinyang and Wang, Fan},
  journal = {arXiv preprint arXiv:2508.17614},
  year    = {2025}
}

@inproceedings{fwgan2019,
  title     = {FW-GAN: Flow-Navigated Warping GAN for Video Virtual Try-On},
  author    = {Dong, Haoye and Liang, Xiaodan and Shen, Xiaohui and Wu, Bowen and Chen, Bing-Cheng and Yin, Jian},
  booktitle = {ICCV},
  year      = {2019}
}

@inproceedings{clothformer2022,
  title     = {ClothFormer: Taming Video Virtual Try-on in All Module},
  author    = {Jiang, Jianbin and Wang, Tan and Yan, He and Liu, Junhui},
  booktitle = {CVPR},
  year      = {2022}
}

@article{vivid2024,
  title   = {ViViD: Video Virtual Try-on using Diffusion Models},
  author  = {Fang, Zixun and Zhai, Wei and Su, Aimin and Song, Hongliang and Zhu, Kai and Wang, Mao and Chen, Yu and Liu, Zhiheng and Cao, Yang and Zha, Zheng-Jun},
  journal = {arXiv preprint arXiv:2405.11794},
  year    = {2024}
}

@inproceedings{tunneltryon2024,
  title     = {Tunnel Try-on: Excavating Spatial-Temporal Tunnels for High-Quality Virtual Try-on in Videos},
  author    = {Xu, Zhengze and Chen, Mengting and Wang, Zhao and Xing, Linyu and Zhai, Zhonghua and Sang, Nong and Lan, Jinsong and Xiao, Shuai and Gao, Changxin},
  booktitle = {ACM Multimedia},
  year      = {2024}
}

@inproceedings{wildvidfit2024,
  title     = {WildVidFit: Video Virtual Try-on in the Wild via Image-based Controlled Diffusion Models},
  author    = {He, Zijian and Chen, Peixin and Wang, Guangrun and Li, Guanbin and Torr, Philip HS and Lin, Liang},
  booktitle = {ECCV},
  year      = {2024}
}

@inproceedings{dpidm2025,
  title     = {Pursuing Temporal-Consistent Video Virtual Try-on via Dynamic Pose Interaction},
  author    = {Li, Dong and Zhong, Wenqi and Yu, Wei and Pan, Yingwei and Zhang, Dingwen and Yao, Ting and Han, Junwei and Mei, Tao},
  booktitle = {CVPR},
  year      = {2025}
}

@article{realvvt2025,
  title   = {RealVVT: Towards Photorealistic Video Virtual Try-on via Spatio-Temporal Consistency},
  author  = {Li, Siqi and Jiang, Zhengkai and Zhou, Jiawei and Liu, Zhihong and Chi, Xiaowei and Wang, Haoqian},
  journal = {arXiv preprint arXiv:2501.08682},
  year    = {2025}
}

@inproceedings{swifttry2025,
  title     = {SwiftTry: Fast and Consistent Video Virtual Try-On with Diffusion Models},
  author    = {Nguyen, Hung and Nguyen, Quang Qui-Vinh and Nguyen, Khoi and Nguyen, Rang},
  booktitle = {AAAI},
  year      = {2025}
}

@article{vitondit2025,
  title   = {VITON-DiT: Learning In-the-Wild Video Try-on from Human Dance Videos via Diffusion Transformers},
  author  = {Zheng, Jun and Zhao, Fuwei and Xu, Youjiang and Dong, Xin and Liang, Xiaodan},
  journal = {arXiv preprint arXiv:2405.18326},
  year    = {2024}
}

@article{catv2ton2025,
  title   = {CatV2TON: Taming Diffusion Transformers for Vision-based Virtual Try-on with Temporal Concatenation},
  author  = {Chong, Zheng and Zhang, Wenqing and Zhang, Shiyue and Zheng, Jun and Dong, Xiao and Li, Haoxiang and Wu, Yiling and Jiang, Dongmei and Liang, Xiaodan},
  journal = {arXiv preprint arXiv:2501.11325},
  year    = {2025}
}

@article{magictryon2025,
  title   = {MagicTryOn: Harnessing Diffusion Transformer for Garment-Preserving Video Virtual Try-on},
  author  = {Li, Guangyuan and Zheng, Siming and Zhang, Hao and Chen, Jinwei and Luan, Junsheng and Ou, Binkai and Zhao, Lei and Li, Bo and Jiang, Peng-Tao},
  journal = {arXiv preprint arXiv:2505.21325},
  year    = {2025}
}

@article{dreamvvt2025,
  title   = {DreamVVT: Mastering Realistic Video Virtual Try-on in the Wild via a Stage-wise Diffusion Transformer Framework},
  author  = {Zuo, Tongchun and Huang, Zaiyu and Ning, Shuliang and Lin, Ente and Liang, Chao and Zheng, Zerong and Jiang, Jianwen and Zhang, Yuan and Gao, Mingyuan and Dong, Xin},
  journal = {arXiv preprint arXiv:2508.02807},
  year    = {2025}
}

@inproceedings{keytailer2026,
  title     = {The Devil is in the Details: Enhancing Video Virtual Try-On via Keyframe-Driven Details Injection},
  author    = {He, Qingdong and Chen, Xueqin and Pan, Yanjie and Tang, Peng and Xu, Pengcheng and Gan, Zhenye and Wang, Chengjie and Hu, Xiaobin and Zhang, Jiangning and Wang, Yabiao},
  booktitle = {CVPR},
  year      = {2026}
}

@article{eevee2025,
  title   = {Eevee: Towards Close-up High-Resolution Video-based Virtual Try-on},
  author  = {Zeng, Jianhao and Bai, Yancheng and Chen, Ruidong and Zhang, Xuanpu and Sun, Lei and Jin, Dongyang and Xu, Ryan and Zhang, Nannan and Song, Dan and Chu, Xiangxiang},
  journal = {arXiv preprint arXiv:2511.18957},
  year    = {2025}
}

@article{itryon2026,
  title   = {iTryOn: Mastering Interactive Video Virtual Try-On with Spatial-Semantic Guidance},
  author  = {Zheng, Jun and Xu, Zhengze and Chen, Mengting and Wang, Jing and Lan, Jinsong and Zhu, Xiaoyong and Zhang, Kaifu and Zheng, Bo and Liang, Xiaodan},
  journal = {arXiv preprint arXiv:2605.21431},
  year    = {2026}
}

@article{pemfvto2025,
  title     = {PEMF-VTO: Point-Enhanced Video Virtual Try-on via Mask-free Paradigm},
  author    = {Chang, Tianyu and Chen, Xiaohao and Wei, Zhichao and Zhang, Xuanpu and Chen, Qingguo and Luo, Weihua and Song, Peipei and Yang, Xun},
  journal   = {IEEE Transactions on Consumer Electronics},
  year      = {2025},
  publisher = {IEEE}
}

@article{tripvvt2025,
  title   = {TripVVT: A Large-Scale Triplet Dataset and a Coarse-Mask Baseline for In-the-Wild Video Virtual Try-On},
  author  = {Shao, Dingbao and Wu, Song and Wang, Shenyi and Wang, Ye and Tang, Ziheng and Liu, Fei and Lin, Jiang and Chen, Xinyu and Wang, Qian and Tai, Ying},
  journal = {arXiv preprint arXiv:2604.27958},
  year    = {2026}
}

@inproceedings{vanast2026,
  title     = {Vanast: Virtual Try-On with Human Image Animation via Synthetic Triplet Supervision},
  author    = {Cha, Hyunsoo and Woo, Wonjung and Kim, Byungjun and Joo, Hanbyul},
  booktitle = {CVPR},
  year      = {2026}
}

@article{ddpm2020,
  title   = {Denoising Diffusion Probabilistic Models},
  author  = {Ho, Jonathan and Jain, Ajay and Abbeel, Pieter},
  journal = {NeurIPS},
  volume  = {33},
  year    = {2020}
}

@inproceedings{ldm2022,
  title     = {High-Resolution Image Synthesis with Latent Diffusion Models},
  author    = {Rombach, Robin and Blattmann, Andreas and Lorenz, Dominik and Esser, Patrick and Ommer, Bj{\"o}rn},
  booktitle = {CVPR},
  year      = {2022}
}

@inproceedings{dit2023,
  title     = {Scalable Diffusion Models with Transformers},
  author    = {Peebles, William and Xie, Saining},
  booktitle = {ICCV},
  year      = {2023}
}

@inproceedings{animatediff2024,
  title     = {AnimateDiff: Animate Your Personalized Text-to-Image Diffusion Models without Specific Tuning},
  author    = {Guo, Yuwei and Yang, Ceyuan and Rao, Anyi and Liang, Zhengyang and Wang, Yaohui and Qiao, Yu and Agrawala, Maneesh and Lin, Dahua and Dai, Bo},
  booktitle = {ICLR},
  year      = {2024}
}

@inproceedings{flowvid2024,
  title     = {FlowVid: Taming Imperfect Optical Flows for Consistent Video-to-Video Synthesis},
  author    = {Liang, Feng and Wu, Bichen and Wang, Jialiang and Yu, Licheng and Li, Kunpeng and Zhao, Yinan and Misra, Ishan and Huang, Jia-Bin and Zhang, Peizhao and Vajda, Peter},
  booktitle = {CVPR},
  year      = {2024}
}

@inproceedings{gowiththeflow2025,
  title     = {Go-with-the-Flow: Motion-Controllable Video Diffusion Models Using Real-Time Warped Noise},
  author    = {Burgert, Ryan and Xu, Yuancheng and Xian, Wenqi and Pilarski, Oliver and Clausen, Pascal and He, Mingming and Ma, Li and Deng, Yitong and Li, Lingxiao and Mousavi, Mohsen},
  booktitle = {CVPR},
  year      = {2025}
}

@inproceedings{flowloss2025,
  title     = {FlowLoss: Dynamic Flow-Conditioned Loss Strategy for Video Diffusion Models},
  author    = {Wu, Kuanting and Ota, Kei and Kanezaki, Asako},
  booktitle = {MVA},
  year      = {2025}
}

@article{moalign2025,
  title   = {MoAlign: Motion-centric Representation Alignment for Video Diffusion Models},
  author  = {Bhowmik, Aritra and Korzhenkov, Denis and Snoek, Cees GM and Habibian, Amirhossein and Ghafoorian, Mohsen},
  journal = {arXiv preprint arXiv:2510.19022},
  year    = {2025}
}

@inproceedings{flowv2v2025,
  title     = {Consistent Video Editing as Flow-Driven Image-to-Video Generation},
  author    = {Wang, Ge and Fan, Songlin and Liu, Hangxu and Song, Quanjian and Wang, Hewei and Xu, Jinfeng},
  booktitle = {CVPR},
  year      = {2026}
}

@inproceedings{motionprompt2025,
  title     = {Optical-Flow Guided Prompt Optimization for Coherent Video Generation},
  author    = {Nam, Hyelin and Kim, Jaemin and Lee, Dohun and Ye, Jong Chul},
  booktitle = {CVPR},
  year      = {2025}
}

@inproceedings{motionagent2025,
  title     = {MotionAgent: Fine-grained Controllable Video Generation via Motion Field Agent},
  author    = {Liao, Xinyao and Zeng, Xianfang and Wang, Liao and Yu, Gang and Lin, Guosheng and Zhang, Chi},
  booktitle = {ICCV},
  year      = {2025}
}

@article{mogan2025,
  title   = {MoGAN: Improving Motion Quality in Video Diffusion via Few-Step Motion Adversarial Post-Training},
  author  = {Xue, Haotian and Chen, Qi and Wang, Zhonghao and Huang, Xun and Shechtman, Eli and Xie, Jinrong and Chen, Yongxin},
  journal = {arXiv preprint arXiv:2511.21592},
  year    = {2025}
}

@inproceedings{onlyflow2025,
  title     = {OnlyFlow: Optical Flow based Motion Conditioning for Video Diffusion Models},
  author    = {Koroglu, Mathis and Caselles-Dupr{\'e}, Hugo and Jeanneret, Guillaume and Cord, Matthieu},
  booktitle = {CVPR},
  year      = {2025}
}

@inproceedings{ropecraft2025,
  title     = {RoPECraft: Training-Free Motion Transfer with Trajectory-Guided RoPE Optimization on Diffusion Transformers},
  author    = {G{\"o}kmen, Ahmet Berke and Ekin, Yi{\u{g}}it and Bilecen, Bahri Batuhan and Dundar, Aysegul},
  booktitle = {NeurIPS},
  year      = {2025}
}

@article{openpose2021,
  title   = {OpenPose: Realtime Multi-Person 2D Pose Estimation using Part Affinity Fields},
  author  = {Cao, Zhe and Hidalgo Martinez, Gines and Simon, Tomas and Wei, Shih-En and Sheikh, Yaser A},
  journal = {IEEE Transactions on Pattern Analysis and Machine Intelligence},
  volume  = {43},
  number  = {1},
  year    = {2021}
}

@inproceedings{densepose2018,
  title     = {DensePose: Dense Human Pose Estimation in the Wild},
  author    = {G{\"u}ler, R{\i}za Alp and Neverova, Natalia and Kokkinos, Iasonas},
  booktitle = {CVPR},
  year      = {2018}
}

@inproceedings{raft2020,
  title     = {RAFT: Recurrent All-Pairs Field Transforms for Optical Flow},
  author    = {Teed, Zachary and Deng, Jia},
  booktitle = {ECCV},
  year      = {2020}
}

@article{ssim,
  title   = {Image Quality Assessment: From Error Visibility to Structural Similarity},
  author  = {Wang, Zhou and Bovik, Alan C and Sheikh, Hamid R and Simoncelli, Eero P},
  journal = {IEEE Transactions on Image Processing},
  volume  = {13},
  number  = {4},
  year    = {2004}
}

@inproceedings{lpips,
  title     = {The Unreasonable Effectiveness of Deep Features as a Perceptual Metric},
  author    = {Zhang, Richard and Isola, Phillip and Efros, Alexei A and Shechtman, Eli and Wang, Oliver},
  booktitle = {CVPR},
  year      = {2018}
}

@inproceedings{vfid,
  title     = {On Aliased Resizing and Surprising Subtleties in GAN Evaluation},
  author    = {Parmar, Gaurav and Zhang, Richard and Zhu, Jun-Yan},
  booktitle = {CVPR},
  year      = {2022}
}

@inproceedings{snr2023,
  title     = {Efficient Diffusion Training via Min-SNR Weighting Strategy},
  author    = {Hang, Tiankai and Gu, Shuyang and Li, Chen and Bao, Jianmin and Chen, Dong and Hu, Han and Geng, Xin and Guo, Baining},
  booktitle = {ICCV},
  year      = {2023}
}

@inproceedings{zerosnr2023,
  title     = {Common Diffusion Noise Schedules and Sample Steps are Flawed},
  author    = {Lin, Shanchuan and Liu, Bingchen and Li, Jiashi and Yang, Xiao},
  booktitle = {WACV},
  year      = {2024}
}

\end{document}